\documentclass[12pt,a4]{article}

\usepackage{adjustbox}
\usepackage{amsmath}
\usepackage{amssymb}
\usepackage{dutchcal}
\usepackage{fancyhdr}
\usepackage{float}
\usepackage[T1]{fontenc}
\usepackage{graphicx}
\usepackage[utf8]{inputenc}
\usepackage{mathtools}
\usepackage{multicol}
\usepackage{multirow}
\usepackage{wasysym}
\usepackage[svgnames]{xcolor}%
\usepackage{xurl}
\usepackage{xcolor}
\usepackage{hyperref}
\usepackage{mdframed}
\definecolor{light-gray}{gray}{0.95}

\newenvironment{note}
{\begin{mdframed}[backgroundcolor=light-gray, roundcorner=10pt,leftmargin=1, rightmargin=1, innerleftmargin=15, innertopmargin=15,innerbottommargin=15, outerlinewidth=1, linecolor=light-gray]}
{\end{mdframed}}

\title{An Introduction to Autoencoders}
\author{Umberto Michelucci\\[3px]\includegraphics[width=2cm]{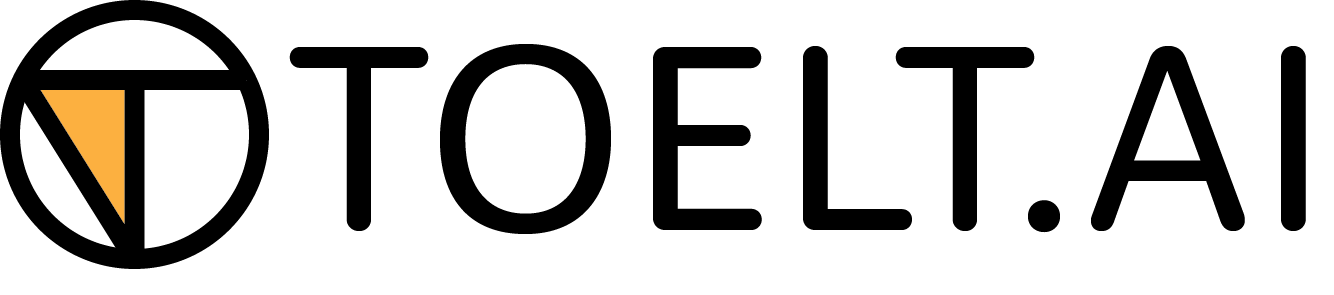}\\[-5px]umberto.michelucci@toelt.ai}

\newtheorem{definition}{Definition}

\begin{document}
\maketitle

In this article, we will look at autoencoders. This article covers the mathematics and the fundamental concepts of autoencoders. We will discuss what they are, what the limitations are, the typical use cases, and we will look at some examples. We will start with a general introduction to autoencoders, and we will discuss the role of the activation function in the output layer and the loss function. We will then discuss what the reconstruction error is. Finally, we will look at typical applications as dimensionality reduction, classification, denoising, and anomaly detection.

\section{Introduction}

Neural networks are typically used in a supervised setting. Meaning that for each training observation \(\mathbf{x}_{i}\)\textbf{ }we will have one label or expected value \(\mathbf{y}_{i}\). During training, the neural network model will learn the relationship between the input data and the expected labels. Now suppose we have only unlabeled observations, meaning we only have our training dataset \( S_{T}\), made of the \( M\) observations \(\mathbf{x}_{i}\) with \( i = 1,\ldots ,M\)
\begin{equation}
S_{T} =\left\{\mathbf{x}_{i} \vert  i = 1,\ldots ,M\right\}   \\ 
\end{equation}

Where in general \(\mathbf{x}_{i}\in\mathbb{R}^{n}\) with \( n\in\mathbb{N}\). Autoencoders were first introduced\footnote{ You can check Rumelhart, D.E., Hinton, G.E. Williams, R.J.: Parallel distributed processing: Explorations in the microstructure of cognition, Vol. 1. Chap. Learning Internal Representations by Error Propagation, pp. 318-162, MIT Press, Cambridge, MA, USA (1986).} by Rumelhart, Hinton, and Williams in 1986 with the goal of learning to reconstruct the input observations \(\mathbf{x}_{i}\) with the lowest error possible\footnote{ In this article we will discuss at length what we mean with error here.}.

Why would one want to learn to reconstruct the input observations? If you have problems imagining what that means, think of having a dataset made of images. An autoencoder would be an algorithm that can give as output an image that is as similar as possible to the input one. You may be confused, as there is no apparent reason of doing so. To better understand why autoencoders are useful we need a more informative (although not yet unambiguous) definition.

\begin{note}
\begin{definition}
An autoencoder is a type of algorithm with the primary purpose of learning an "informative" representation of the data that can be used for different applications\footnote{ Bank, D., Koenigstein, N., and Giryes, R., Autoencoders, \url{https://arxiv.org/abs/2003.05991} } by learning to reconstruct a set of input observations well enough.
\end{definition}
\end{note}

\begin{figure}[hbt]
\centering
\includegraphics[width=12.61cm,height=3.97cm]{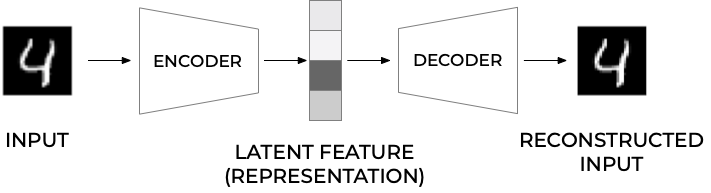}
\caption{General structure of an autoencoder.}
\label{fig:arch}
\end{figure}

To better understand autoencoders we need to refer to their typical architecture, visualized in Figure \ref{fig:arch}. The autoencoders' main components are three: an encoder, a latent feature representation, and a decoder. The encoder and decoder are simply functions, while with the name \textit{latent feature representation,} one typically intends a tensor of real numbers (more on that later). Generally speaking, we want the autoencoder to reconstruct the input well enough. Still, at the same time, it should create a latent representation (the output of the \textbf{encoder} part in Figure \ref{fig:arch}) that is useful and meaningful. For example, latent features on hand-written digits\footnote{ Consider for example the MNIST dataset: \url{http://yann.lecun.com/exdb/mnist/}} could be the number of lines required to write each number or the angle of each line and how they connect. Learning how to write numbers certainly does not require to learn the gray values of each pixel in the input image. We humans do not certainly learn to write by filling pixels with gray values. While learning, we extract the essential information that will allow us to solve a problem (writing digits, for example). This latent representation (\textit{how} to write each number) can then be very useful for various tasks (for instance feature extraction that can be then used for classification or clustering) or simply understanding the essential features of a dataset.

In most typical architectures, the encoder and the decoder are neural networks\footnote{ For example, if the encoder and the decoder are linear functions, you get what is called a linear autoencoder. See for more information Baldi, P., Hornik, K.: Neural networks and principal component analysis: Learning from examples without local minima, Neural Netw. 2(1), 53-58 (1989).} (that is the case we will discuss at length in this article) since they can be easily trained with existing software libraries such as TensorFlow or PyTorch with backpropagation. 
In general, the encoder can be written as a function \( g\) that will depend on some parameters
\begin{equation}
\mathbf{h}_{i} = g(\mathbf{x}_{i})
\end{equation}
Where \(\mathbf{h}_{i}\in\mathbb{R}^{q}\) (the latent feature representation) is the output of the \textbf{encoder} block in Figure \ref{fig:arch} when we evaluate it on the input \(\mathbf{x}_{i}\). Note that we will have \( g:\mathbb{R}^{n}\rightarrow\mathbb{R}^{q}\).
The \textbf{decoder} (and the output of the network that we will indicate with \(\tilde{\mathbf{x}}_{i}\)) can be written then as a second generic function \( f\) of the latent features
\begin{equation}
\tilde{\mathbf{x}}_{i} = f\left(\mathbf{h}_{i}\right) = f\left(g\left(\mathbf{x}_{i}\right)\right).
\end{equation}
Where \(\tilde{\mathbf{x}}_{i}\mathbf{\in }\mathbb{R}^{n}\). Training an autoencoder simply means finding the functions \( g(\cdot)\) and \( f(\cdot)\) that satisfy
\begin{equation}
\textrm{arg}\min_{f,g}<\left[\Delta (\mathbf{x}_{i}, f(g\left(\mathbf{x}_{i}\right))\right]>
\end{equation}
where \( \Delta\) indicates a measure of how the input and the output of the autoencoder differ (basically our loss function will penalize the difference between input and output) and \( <\cdot>\) indicates the average over all observations.  Depending on how one designs the autoencoder, it may be possible to find \( f\) and \( g\) so that the autoencoder learns to reconstruct the output perfectly, thus learning the identity function. This is not very useful, as we discussed at the beginning of the article, and to avoid this possibility, two main strategies can be used: creating a bottleneck and add regularization in some form. 
\begin{note}
\textbf{Note} We want the autoencoder to reconstruct the input well enough. Still, at the same time, it should create a latent representation (the output of the encoder) that is useful and meaningful.
\end{note}
Adding a "bottleneck," (more on that later) is achieved by making the latent feature's dimensionality lower (often much lower) than the input's. That is the case that we will look in detail in this article. But before looking at this case, let’s briefly discuss regularization.

\subsection{Regularization in autoencoders}

We will not discuss regularization at length in this article, but we should at least mention it. Intuitively it means enforcing sparsity in the latent feature output. The simplest way of achieving this is to add a \(\mathcal{l}_{1}\) or \(\mathcal{l}_{2}\) regularization term to the loss function. That will look like this for the \(\mathcal{l}_{2}\) regularization term:
\begin{equation}
\textrm{arg}\min_{f,g}\mathbb{(E}\left[\Delta (\mathbf{x}_{i}, g(f\left(\mathbf{x}_{i}\right))\right]+\lambda \sum_{i}^{}\theta_{i}^{2})
\end{equation}
In the formula the \( \theta_{i}\) are the parameters in the functions \( f(\cdot)\) and \( g(\cdot)\) (you can imagine that in the case where the functions are neural networks, the parameters will be the weights). This is typically easy to implement as the derivative with respect to the parameters are easy to calculate. Another trick that is worth mentioning is to tie the weights of the encoder to the weights of the decoder\footnote{ For an example in Keras you can check the following page: \url{http://adl.toelt.ai/Chapter25/Your_first_autoencoder_with_Keras.html} } (in other words make them equal). Those techniques, and a few others that go beyond the scope of this book, have fundamentally the same effect: add sparsity to the latent feature representation.

We turn now to a specific type of autoencoders: those that build \( f\) and \( g\) with feed-forward networks that use a bottleneck. The reason for this choice is that they are very easy to implement and are very effective.

\section{Feed Forward Autoencoders}

\begin{figure}[hbt]
\label{fig:arch2}
\centering
\includegraphics[width=12.6cm,height=7.85cm]{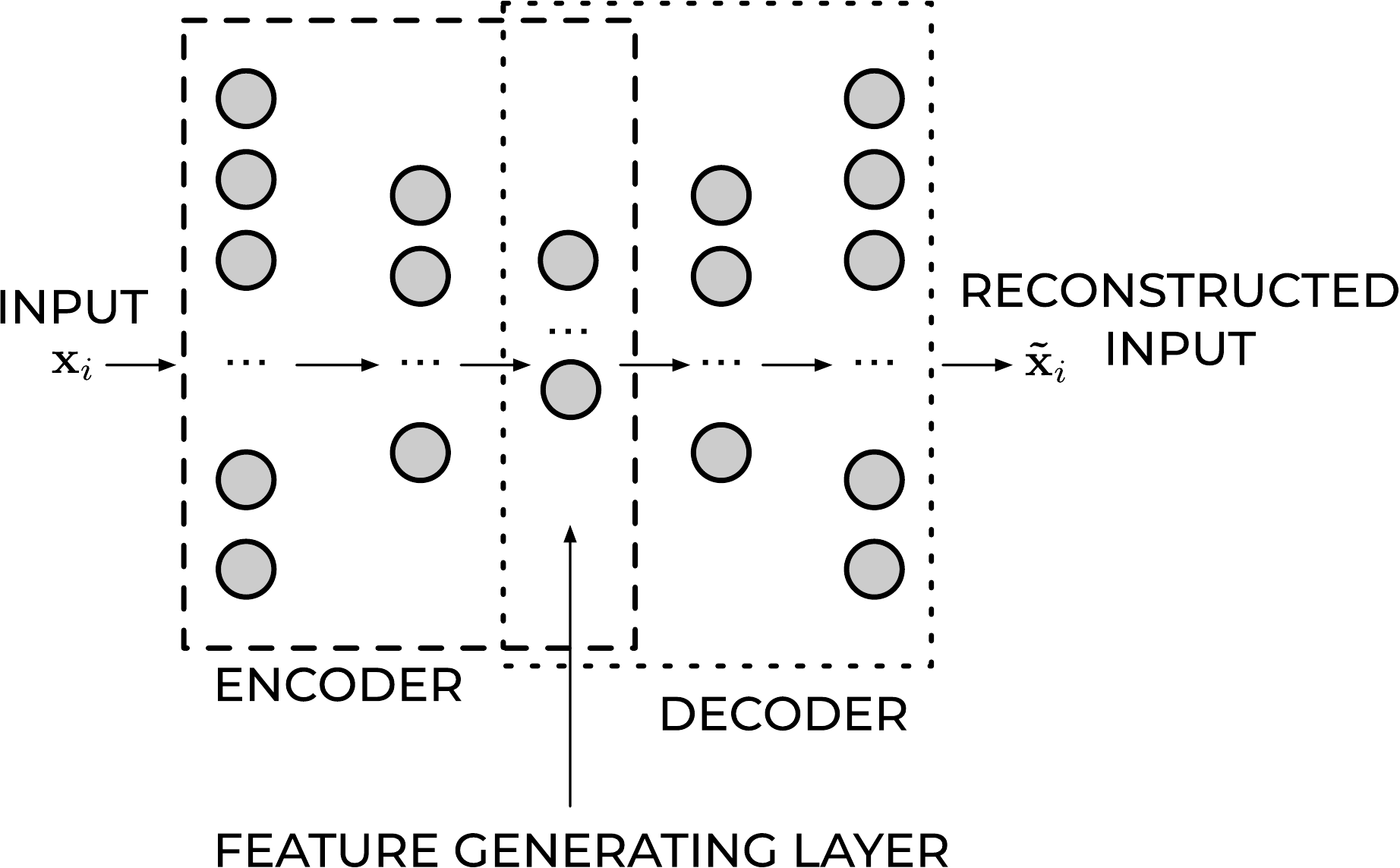}
\caption{A typical architecture of a Feed-Forward Autoencoder. The number of neurons in the layers at first goes down as we move through the network until it reaches the middle and then starts to grow again until the last layer has the same number of neurons as the input dimensions.}
\end{figure}

A Feed-Forward Autoencoder (FFA) is a neural network made of dense layers\footnote{ A dense layer is simply a set of neurons that gets their inputs from the previous layer. Each neuron in a dense layer gets as input the output of all neurons in the previous layer.} with a specific architecture, as can be schematically seen in Figure \ref{fig:arch2}.

A typical FFA architecture (although it is no mandatory requirement) has an odd number of layers and is symmetrical with respect to the middle layer. Typically, the first layer has a number of neurons \( n_{1} = n\) (the size of the input observation \(\mathbf{x}_{\mathbf{i}}\)). As we move toward the center of the network, the number of neurons in each layer drops in some measure. The middle layer (remember we have an odd number of layers) usually has the smallest number of neurons. The fact that the number of neurons in this layer is smaller than the size of the input, is the \textbf{bottleneck} we mentioned earlier.

In almost all practical applications, the layers after the middle one are a mirrored version of the layers before the middle one. For example, an autoencoder with 3 layers could have the following numbers of neurons: \( n_{1} = 10\), \( n_{2} = 5\) and then \( n_{3} = n_{1} = 10\) (supposing we are working on a problem where the input dimension is \( n = 10\)). All the layers up to and including the middle one, make what is called the encoder, and all the layers from and including the middle one (up to the output) make what is called the decoder, as you can see depicted in Figure (25.2). If the FFA training is successful, the result will be a good approximation of the input, in other words \(\tilde{\mathbf{x}}_{i}\approx\mathbf{x}_{i}\). What is essential to notice is that the decoder can reconstruct the input by using only a much smaller number (\( q\)) of features than the input observations initially have (\( n\)). The output of the middle layer \(\mathbf{h}_{\mathbf{i}}\)\textbf{ }are also called a \textit{learned representation} of the input observation \(\mathbf{x}_{i}\). 

\begin{note}
{\bf Note} The encoder can reduce the number of dimensions of the input observation (\( n\)) and create a learned representation (\(\mathbf{h}_{\mathbf{i}}\mathbf{) }\)of the input that has a smaller dimension \( q<n\). This learned representation is enough for the decoder to reconstruct the input accurately (if the autoencoder training was successful as intended). 
\end{note}

\subsection{Activation Function of the Output Layer}

In autoencoders based on neural networks, the output layer's activation function plays a particularly important role.  The most used functions are ReLU and sigmoid. Let us look at both and look at some tips on when to use which and why you should choose one instead of the other.

\subsubsection{ReLU}

The  ReLU activation function can assume all values in the range \(\left[0,\infty\right]\). As a remainder, its formula is
\begin{equation}
\textrm{ReLU}\left(x\right) = \max\left(0,x\right).
\end{equation}
This choice is good when the input observations \(\mathbf{x}_{i}\) assume a wide range of positive values.
If the input \(\mathbf{x}_{i}\) can assume negative values, the ReLU is, of course, a terrible choice, and the identity function is a much better choice.
\begin{note}
\textbf{Note }ReLU activation function for the output layer is well suited for cases when the input observations \(\mathbf{x}_{i}\) assume a wide range of positive real values.
\end{note}
\subsubsection{Sigmoid}

The sigmoid function \( \sigma\) can assume all values in the range \( ]0,1[\). As a remained its formula is

\begin{equation}
\sigma\left(x\right) =\frac{1}{1+e^{-x}}.
\end{equation}

This activation function can only be used if the input observations \(\mathbf{x}_{i}\) are all in the range \( ]0,1[\) or if you have normalized them to be in that range. Consider as an example the MNIST dataset. Each value of the input observation \(\mathbf{x}_{i}\) (one image) is the gray values of the pixels that can assume any value from 0 to  255. Normalizing the data by dividing the pixel values by 255 would make each observation (each image) have only pixel values between 0 and 1. In this case, the sigmoid would be a good choice for the output layer's activation function. 

\begin{note}
\textbf{Note }The sigmoid activation function for the output layer is a good choice in all cases where the input observations assume only values between \( 0\) and \( 1\) or if you have normalized them to assume values in the range \( ]0,1[\).
\end{note}

\subsection{Loss Function}

As with any neural network model, we need a loss function to minimize. This loss functions should measure how big is the difference between the input \(\mathbf{x}_{i}\) and output \(\tilde{\mathbf{x}}_{i}\). If you remember the explanations at the beginning, you will realize that our loss function will be

\begin{equation}
\mathbb{E}\left[\Delta (\mathbf{x}_{i}, g(f\left(\mathbf{x}_{i}\right))\right].
\end{equation}

Where for FFAs, \( g,\) and \( f\) will be the functions that are obtained with dense layers, as discussed in the previous sections. Remember that an autoencoder is trying to learn an approximation of the identity function; therefore, you want to find the weights in the network that gives you the smallest difference according to some metric (\( \Delta (\cdot)\)) between \(\mathbf{x}_{i}\) and \(\tilde{\mathbf{x}}_{i}\). Two loss functions are widely used for autoencoders:  Mean Squared Error (MSE) and Binary Cross-Entropy (BCE). Let us have a more in-depth look at both since they can only be used when specific requirements are met.

\subsubsection{Mean Square Error}

Since an autoencoder is trying to solve a regression problem, the most common choice as a loss function is the Mean Square Error (MSE):
\begin{equation}
\begin{split}
L_{\textrm{MSE}} = \textrm{MSE} = \frac{1}{M}\sum_{i = 1}^{M}\left\vert\mathbf{x}_{i}-\tilde{\mathbf{x}}_{i}\right\vert^{2} \\ 
\end{split}
\end{equation}
The symbol \( \vert \cdot\vert\) indicates the norm of a vector\footnote{ The norm of a vector is simply the square root of the sum of the square of the components.}, and M is the number of the observation in the training dataset. It can be used in almost all cases, independently of how you choose your output layer activation function or how you normalize the input data. 
It is easy to show that the minimum of \( L_{\textrm{MSE}}\) is found for \(\tilde{\mathbf{x}}_{i} =\mathbf{x}_{i}\). To prove it, let us calculate the derivative of \( L_{\textrm{MSE}}\) with respect to a specific observation \( j.\) Remember that the minimum is found when the condition
\begin{equation}
\begin{split}
\frac{\partial L_{MSE}}{\partial\tilde{x}_{j}} = 0 \\ 
\end{split}
\end{equation}
is met for all \( i = 1,\ldots ,M\). To simplify the calculations, let us assume that the inputs are one dimensional\footnote{ If we did not make this assumption, one would have to calculate the gradient of the loss function instead of the simple derivative.} and let us indicate them with \( x_{i}\). We can write
\begin{equation}
\begin{split}
\frac{\partial L_{MSE}}{\partial\tilde{x}_{j}} =  -\frac{2}{M}\left(x_{j}-\tilde{x}_{j}\right) \\ 
\end{split}
\end{equation}
Equation \ref{fig:arch} is satisfied when \( x_{j} =\tilde{x}_{j}\) as can be easily seen from Equation (25.1), as we wanted to prove. To be precise, we also need to show that
\begin{equation}
\begin{split}
\frac{\partial^{2}L_{MSE}}{\partial\tilde{x}_{j}^{2}}>0 \\ 
\end{split}
\end{equation}
This is easily proved as we have
\begin{equation}
\begin{split}
\frac{\partial^{2}L_{MSE}}{\partial\tilde{x}_{j}^{2}} =\frac{2}{M} \\ 
\end{split}
\end{equation}
sthat is greater than zero, therefore confirming our assumption that for \( x_{j} =\tilde{x}_{j}\) we indeed have a minimum.

\subsubsection{Binary Cross-Entropy}

If the activation function of the output layer of the FFA is a sigmoid function, thus limiting neuron outputs to be between 0 and 1, and the input features are normalized to be between 0 and 1 we can use as loss function the binary cross-entropy, indicated here with \( L_{\textrm{CE}}\). Note that this loss function is typically used in classification problems, but it works beautifully for autoencoders. The formula for it is
\begin{equation}
\begin{split}
L_{\textrm{CE}} = -\frac{1}{M}\sum_{i = 1}^{M}\sum_{j = 1}^{n}[x_{j,i} \log\tilde{x}_{j,i}+\left(1-x_{j,i}\right)\log (1-\tilde{x}_{j,i})] \\ 
\end{split}
\end{equation}
Where \( x_{j,i}\) is the \( j^{th}\) component of the \( i^{th}\) observation. The sum is over the entire set of observations and over all components of the vectors. Can we prove that minimizing this loss function is equivalent to reconstructing the input as well as possible? Let us calculate where \( L_{\textrm{CE}}\) has a minimum with respect to \(\tilde{\mathbf{x}}_{i}\). In other words, we need to find out what values should \(\tilde{\mathbf{x}}_{i}\) assume to minimize \( L_{\textrm{CE}}\). As we have done for the MSE, to make the calculations easier, let us consider the simplified case where \(\mathbf{x}_{i}\) and \(\tilde{\mathbf{x}}_{i}\) are one-dimensional and let us indicate them with \( x_{i}\) and \(\tilde{x}_{i}\).

To find the minimum of a function, as you should know from calculus, we need the first derivative of \( L_{\textrm{CE}}\). In particular we need to solve the set of \( M\) equations
\begin{equation}
\frac{\partial L_{\textrm{CE}}}{\partial\tilde{x}_{i}} = 0   \ \ \ \text{for}\ \ \       i = 1,\ldots ,M \\ 
\end{equation}
In this case it is easy to show that the binary cross-entropy \( L_{\textrm{CE}}\) is minimized when \( x_{i} =\tilde{x}_{i}\) for \( i = 1,\ldots ,M\). Note that strictly speaking, this is true only for \( x_{i}\)different than 0 or 1 since \(\tilde{x}_{i}\) can be neither 0 nor 1.
To find when the \( L_{CE}\) is minimized we can derive \( L_{CE}\)with respect to a specific input \(\tilde{x}_{j}\)
\begin{equation}
\begin{array}{lll}
\displaystyle \frac{\partial L_{CE}}{\partial\tilde{x}_{j}} &=& -\displaystyle\frac{1}{M}\left[\frac{x_{j}}{\tilde{x}_{j}}-\frac{1-x_{j}}{1-\tilde{x}_{j}}\right] = -\frac{1}{M}\left[\frac{x_{j}\left(1-\tilde{x}_{j}\right)-\tilde{x}_{j}\left(1-x_{j}\right)}{\tilde{x}_{j}-\tilde{x}_{j}^{2}}\right] =\\ [12px]
&= &-\displaystyle\frac{1}{M}\left[\frac{x_{j}-\tilde{x}_{j}}{\tilde{x}_{j}-\tilde{x}_{j}^{2}}\right]
\end{array}
\end{equation}
Now remember that we need to satisfy the condition 
\begin{equation}
\begin{split}
\frac{\partial L_{CE}}{\partial\tilde{x}_{j}} = 0 \\ 
\end{split}
\end{equation}
That can happen only if \( x_{j} =\tilde{x}_{j}\) as can be seen from Equation (25.2). To make sure that this is a minimum we need to evaluate the second derivative. Since the point for which the first derivative is zero is a minimum only if 
\begin{equation}
\begin{split}
\frac{\partial^{2}L_{CE}}{\partial\tilde{x}_{j}^{2}}>0 \\ 
\end{split}
\end{equation}
We can calculate the second derivative at the minimum point \( x_{j} =\tilde{x}_{j}\)easily 
\begin{equation}
\begin{split}
\left.\frac{\partial^{2}L_{CE}}{\partial\tilde{x}_{j}^{2}}\right\vert_{x_{j} =\tilde{x}_{j}} = -\left.\frac{1}{M}\left[\frac{x_{j}\left(2\tilde{x}_{j}-1\right)-\tilde{x}_{j}^{2}}{\left(1-\tilde{x}_{j}^{2}\right)\tilde{x}_{j}^{2}}\right]\right\vert_{x_{j} =\tilde{x}_{j}} =\frac{1}{M}\left[\frac{\tilde{x}_{j}\left(1-\tilde{x}_{j}\right)}{\left(1-\tilde{x}_{j}^{2}\right)\tilde{x}_{j}^{2}}\right] \\ 
\end{split}
\end{equation}
Now remember that \(\tilde{x}_{j}\in ]0,1[\). We can immediately see that the denominator of the previous formula is greater than zero. The nominator is also clearly greater than zero since \( 1-\tilde{x}_{i}>0\). Dividing two positive numbers gives a positive number, thus we have just proved that 
\begin{equation}
\begin{split}
\frac{\partial^{2}L_{CE}}{\partial\tilde{x}_{i}^{2}}>0 \\ 
\end{split}
\end{equation}
The minimum of the cost function is reached when the output is exactly equal to the inputs, as we wanted to prove. 
\begin{note}
\textbf{Note} an essential prerequisite for using the binary cross-entropy loss function is that the inputs \textbf{must} be normalized between 0 and 1 and that the activation function for the last layer must be a \textit{sigmoid} or \textit{softmax} function.
\end{note}
\subsection{Reconstruction Error}

The reconstruction error (RE) is a metric that gives you an indication of how good (or bad) the autoencoder was able to reconstruct the input observation \(\mathbf{x}_{i}\). The most typical RE used is the MSE
\begin{equation}
\begin{split}
\textrm{RE}\equiv \textrm{MSE} = \frac{1}{M}\sum_{i = 1}^{M}\left\vert\mathbf{x}_{i}-\tilde{\mathbf{x}}_{i}\right\vert^{2}\\ 
\end{split}
\end{equation}

\begin{figure}[hbt]
\centering
\includegraphics[width=10.12cm,height=9.16cm]{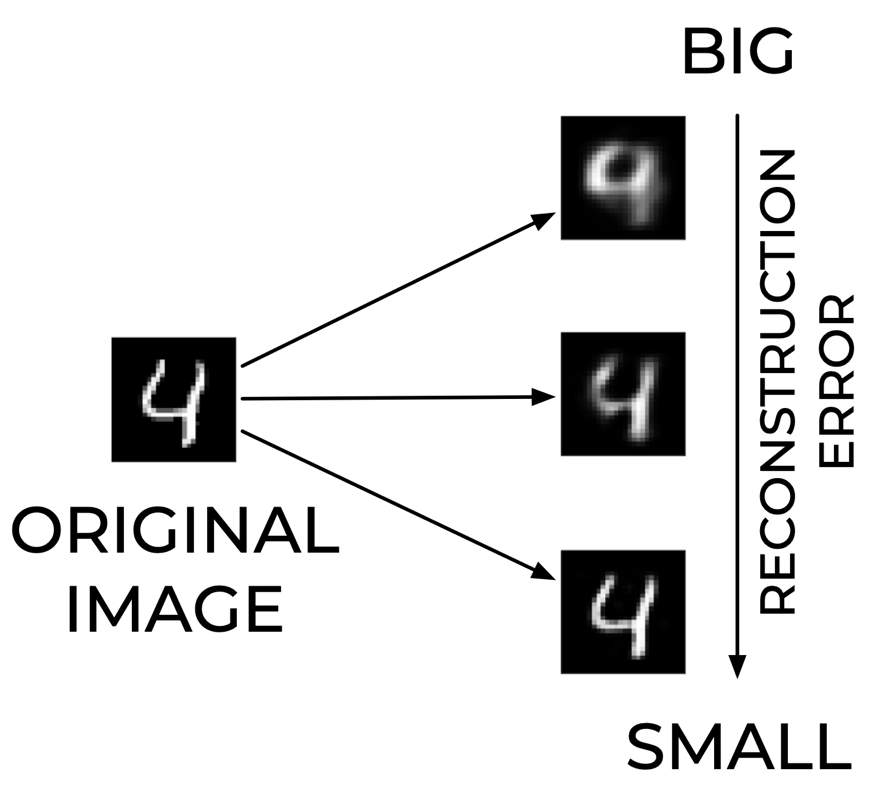}
\caption{An example of big and small reconstruction error when an autoencoder tries to reconstruct an image.}\label{fig:rec_err}
\end{figure}

That can be easily calculated. The RE is used often when doing anomaly detection with autoencoders, as we will explain later. There is an easy intuitive explanation of the reconstruction error. When the RE is significant, the autoencoder could not reconstruct the input well, while when it is small, the reconstruction was successful. Figure \ref{fig:rec_err} shows an example of big and small reconstruction errors when an autoencoder tries to reconstruct an image.

\subsection{Example: reconstruction of hand-written digits}

\begin{figure}[hbt]
\centering
\includegraphics[width=12.61cm,height=2.92cm]{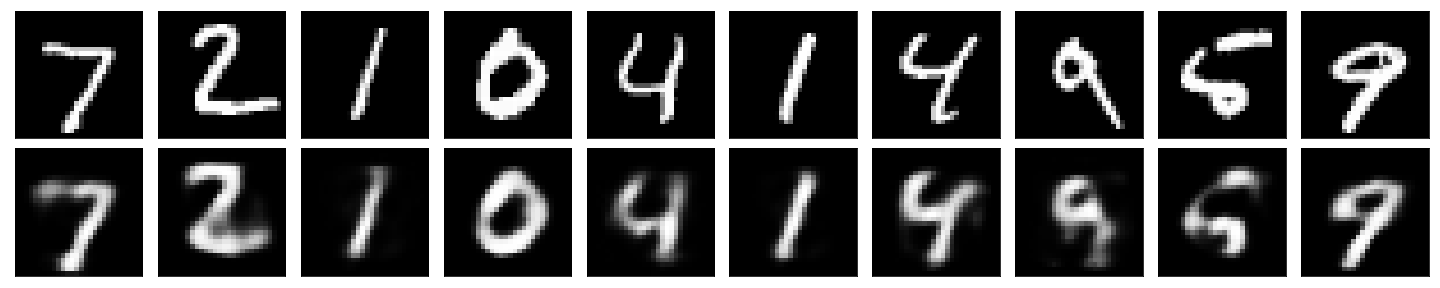}
\caption{In the top line, you can see the original digits from the MNIST dataset. In contrast, the line below contains the digits reconstructed by the autoencoder with number of neurons equal to (784, 16, 784).
}\label{fig:rec_2}
\end{figure}

Let us now see how an autoencoder performs with a real example, using the MNIST dataset. This dataset\footnote{ More information on the dataset can be found here: \url{http://yann.lecun.com/exdb/mnist/}. } contains 70000 hand-written digits from 0 to 9. Each image is \( 28\times 28\) pixels with only gray values, that means that we have 784 features (the pixel gray values) as inputs. Let us start with an autoencoder with 3 layers with the numbers of neurons in each layer equal to \(\left(784,16,784\right)\). Note that the first and last layers must have a dimension equal to the input dimensions. For this example, we used the Adam optimizer\footnote{ You can find the entire code at the address \url{https://adl.toelt.ai}. }, as loss function the cross-entropy\footnote{ In this case we normalized the input features to be between 0 and 1.} and we trained the model for \( 30\) epochs with a batch size of \( 256\). In Figure  \ref{fig:rec_2} you can see two lines of images of digits. The line at the top contains ten random images from the original dataset, while the ones at the bottom are the reconstructed images with the autoencoder we just described.

\begin{figure}[H]
\centering
\includegraphics[width=12.61cm,height=2.92cm]{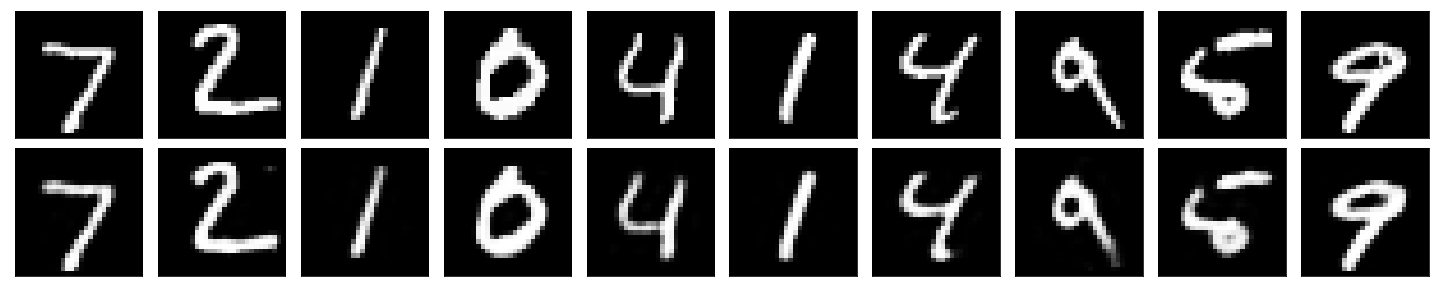}
\caption{n the top line you can see the original digits from the MNIST dataset. While the line below are the digits reconstructed by the autoencoder with number of neurons equal to (784, 64, 784).}\label{fig:rec_3}
\end{figure}

\begin{figure}[hbt]
\centering
\includegraphics[width=12.61cm,height=2.92cm]{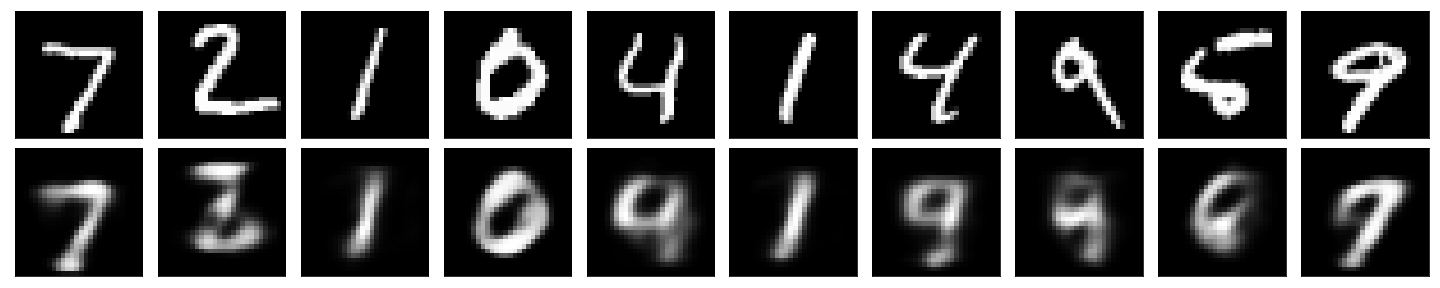}
\caption{In the top line you can see the original digits from the MNIST dataset. In contrast, the line below contains the digits reconstructed by the autoencoder with number of neurons equal to (784, 8, 784).
}\label{fig:rec_4}
\end{figure}

\begin{figure}[hbt]
\centering
\includegraphics[width=12.61cm,height=4.36cm]{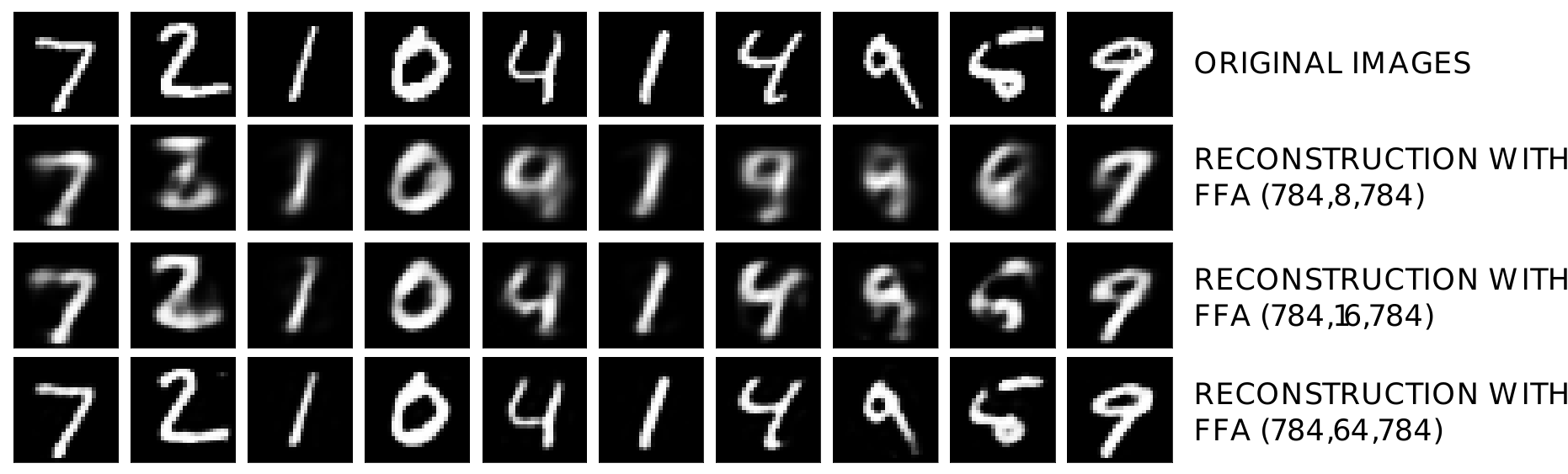}
\caption{In the top line, you can see the original digits from the MNIST dataset. The second line of digits contains the digits reconsructed by the FFA (784,8,784), the third by the FFA (784,16,784), and the last one by the FFA (784,64,784).}\label{fig:rec_5}
\end{figure}

It is impressive that to reconstruct an image with 784 pixels, 10 classes and 70000 images only 16 features are needed to have a result that, although not perfect, allows us to understand almost entirely what digit was used as input. Increasing the middle layer's size to \( 64\) (and leaving all other parameters the same) gets a much better result as you can see in Figure \ref{fig:rec_3}.
This tells us that the relevant information on how to write digits is contained in a much lower number of features than 784. 
\begin{note}
\textbf{Note }An autoencoder with a middle layer smaller than the input dimensions (a bottleneck) can be used to extract the essential features of an input dataset creating a learned representation of the inputs given by the function \( g\left(\mathbf{x}_{i}\right)\). Effectively an FFA can be used to perform dimensionality reduction.
\end{note}
The FFA will not recreate the input digits well if the number of neurons in the middle layer is reduced too much (if the bottleneck is too extreme). Figure \ref{fig:rec_4} shows the reconstruction of the same digits with an autoencoder with only 8 neurons in the middle layer. With only 8 neurons in the middle layer, you can see that some reconstructed digits are wrong. As you can see in Figure \ref{fig:rec_4} the 4 is reconstructed as a 9 and a 2 is reconstructed to something that resembles a 3.
In Figure \ref{fig:rec_5}, you can compare the reconstructed digits by all the FFAs we have discussed.
From Figure \ref{fig:rec_5} you can see how, increasing the middle layer's size, the reconstruction gets better and better, as we expected.

\begin{figure}[hbt]
\label{fig:rec_6}
\centering
\includegraphics[width=12.6cm]{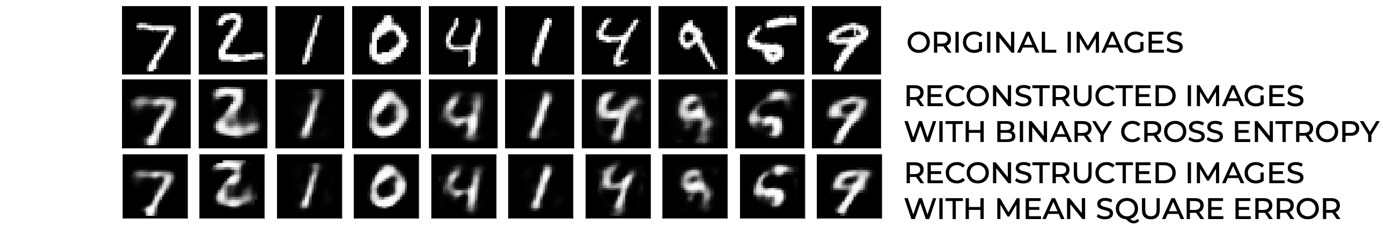}
\caption{In the top line, you can see ten random original digits from the MNIST dataset. The second line of digits contains the digits reconstructed with an FFA with 16 neurons in the middle layer and the binary cross-entropy as the loss function. The last line contains images reconstructed with the MSE as loss function.}
\end{figure}

For these examples, we have used the binary cross entropy as loss function, but the MSE would have worked also well, and results can be seen in Figure \ref{fig:rec_6}.

\section{Autoencoders Applications}

\subsection{Dimensionality Reduction}

As we mentioned in this article, using the bottleneck method, the latent features will have a dimension \( q\) that is smaller than the dimensions of the input observations \( n\). The \textit{encoder} part (once trained) does naturally (by design) dimension reduction producing \( q\) real numbers. One can use the latent features for various tasks, such as classification (as we will see in the next section) or clustering. We would like to point out some of the advantages of dimensionality reduction with an autoencoder compared to a more classical PCA approach. The autoencoder has one main benefit from a computational point of view: it can deal with a very big amount of data efficiently since its training can be done with mini-batches, while PCA, one of the most used dimensionality reduction algorithms, needs to do its calculations using the entire dataset. PCA is an algorithm that projects a dataset on the eigenvectors of its covariance matrix\footnote{ Akshay Balsubramani, Sanjoy Dasgupta, and Yoav Freund. The fast convergence of incremental pca. In Advances in neural information processing systems, pages 3174–3182, 2013.}, thus providing a linear transformation of the features. Autoencoders are more flexible and consider non-linear transformations of the features. The default PCA method uses \(\mathcal{O}\left(d^{2}\right)\) space for data in \(\mathbb{R}^{d}\). This is, in many cases, not computationally feasible, and the algorithm does not scale up with increasing dataset size. This may seem irrelevant, but in many practical applications, the amount of data and the number of features is so big that PCA is not a practical solution from a computational point of view. 
\begin{note}
\textbf{Note }The use of an autoencoder for dimensionality reduction has one main advantage from a computational point of view: it can deal with a very big amount of data efficiently since its training can be done with mini-batches.
\end{note}
\paragraph{Equivalence with PCA}

It is not a very known results, but one worth to mention, that a FFA is equivalent to PCA if the following conditions are met:
\begin{itemize}
\item We use a linear function for the encoder \( g(\cdot)\)
\item We use a linear function for the decoder \( f(\cdot)\)
\item We use the MSE for the loss function
\item We normalize the inputs to 
\begin{equation}
\hat{x}_{i,j} =\frac{1}{\sqrt{M}}\left(x_{i,j}-\frac{1}{M}\sum_{k = 1}^{M}x_{k,j}\right)
\end{equation}
\end{itemize}

The proof is long and can found in the notes by M.M. Kahpra for the course CS7015 (Indian Institute of Technology Madras) at this link \url{http://toe.lt/1a}. 

\subsection{Classification}

\subsubsection{Classification with Latent Features}

Let us now suppose that we want to classify our input images of the MNIST dataset. We can simply use all the features, in our case, the \( 784\) pixel values of the images. We can simply use an algorithm as kNN for illustrative purposes. Doing it with \( 7\) nearest neighbors on the training MNIST dataset (with 60000 images) will take around 16.6 minutes\footnote{ The examples have been run on Google Colab.} (ca. 1000 sec) and gets you an accuracy on the test dataset of 10000 images of \( 96.4 \%\footnote{ Note that for these examples the accuracy is calculated applying the trained model on the test dataset, while the running time is the time needed to train the algorithm on the training dataset.}\). However, what happens if we use this algorithm not with the original dataset, but with the latent features \( g\left(\mathbf{x}_{i}\right)\)? For example, if we consider an FFA with \( 8\) neurons in the middle layer and again train a kNN algorithm on the latent features \( g\left(\mathbf{x}_{i}\right)\in\mathbb{R}^{8}\) we get an accuracy of 89$\%$ in 1.1 sec. We get a gain of a factor of 1000 in running time, for a loss of 7.4$\%$ in accuracy\footnote{ You can run those tests yourself going to article 25 at https://adl.toelt.ai. } (see Table \ref{tab:run1}). 

\begin{table}[hbt]
\begin{adjustbox}{max width=\textwidth}
\begin{tabular}{p{4.89cm}p{4cm}p{4.59cm}}
\multicolumn{1}{p{4.9cm}}{Input Data} & 
\multicolumn{1}{p{4cm}}{Accuracy} & 
\multicolumn{1}{p{4.6cm}}{Running Time} \\ 
\hline
\multicolumn{1}{p{4.9cm}}{Original data \(\mathbf{x}_{i}\in\mathbb{R}^{784}\)} & 
\multicolumn{1}{p{4cm}}{96.4$\%$} & 
\multicolumn{1}{p{4.6cm}}{\( 1000\) sec. \( \approx 16.6\) min.} \\ 
\multicolumn{1}{p{4.9cm}}{Latent Features g\(\left(\mathbf{x}_{i}\right)\in\mathbb{R}^{8}\)} & 
\multicolumn{1}{p{4cm}}{89$\%$} & 
\multicolumn{1}{p{4.6cm}}{1.1 sec.} \\ 
\end{tabular}
\end{adjustbox}
\caption{the different in accuracy and running time when applying the kNN algorithm to the original 784 features or the 8 latent features for the MNIST dataset.}\label{tab:run1}
\end{table}

Using \( 8\) features allow us to get a very high accuracy in just one second. 
We can do the same analysis with another dataset, the Fashion MNIST\footnote{ https://research.zalando.com/welcome/mission/research-projects/fashion-mnist/ } dataset (a dataset from Zalando very similar to the MNIST one, only with clothing images instead of hand-written digits) for illustrative purposes. The dataset has, as the MNIST one, 60000 training images and 10000 test ones. In Table \ref{tab:run2} you can see the summary of the results of applying kNN to the testing portion of this dataset.
\begin{table}[hbt]
\begin{adjustbox}{max width=\textwidth}
\begin{tabular}{p{4.89cm}p{4cm}p{4.59cm}}
\multicolumn{1}{p{4.9cm}}{Input Data} & 
\multicolumn{1}{p{4cm}}{Accuracy} & 
\multicolumn{1}{p{4.6cm}}{Running Time} \\ 
\hline
\multicolumn{1}{p{4.9cm}}{Original data\( \mathbf{x}_{i}\in\mathbb{R}^{784}\)} & 
\multicolumn{1}{p{4cm}}{85.4$\%$} & 
\multicolumn{1}{p{4.6cm}}{\( 1040\) sec. \( \approx 16.6\) min.} \\ 
\multicolumn{1}{p{4.9cm}}{Latent Features \( enc\left(\mathbf{x}_{i}\right)\in\mathbb{R}^{8}\)} & 
\multicolumn{1}{p{4cm}}{79.9$\%$} & 
\multicolumn{1}{p{4.6cm}}{1.2 sec.} \\ 
\multicolumn{1}{p{4.9cm}}{Latent Features \( enc\left(\mathbf{x}_{i}\right)\in\mathbb{R}^{16}\)} & 
\multicolumn{1}{p{4cm}}{83.6$\%$} & 
\multicolumn{1}{p{4.6cm}}{3.0 sec.} \\ 
\end{tabular}
\end{adjustbox}
\caption{the difference in accuracy and running time when applying the kNN algorithm to the original 784 features with a FFA with 8 neurons and with a FFA with 16 neurons for the Fashion MNIST dataset.}\label{tab:run2}
\end{table}
It is exciting to note that with an FFA with 16 neurons in the middle layer, we reach an accuracy of 83.6$\%$ in just 3 sec. When applying a kNN algorithm to the original features (784), we get an accuracy only 1.8$\%$ higher but with a running time ca. 330 times longer.
\begin{note}
\textbf{Note} Using autoencoders and doing classification with the latent features is a very viable technique to reduce the training time by several order of magnitude while incurring a minor drop in accuracy.
\end{note}
\subsubsection{Curse of dimensionality – a small detour}

Is there any other reason why we want to do dimensionality reduction before doing any classification? Reducing running time is one reason, but another important one plays a significant role when the input dimension is very large, i.e., the datasets that have a very high number of features: the curse of dimensionality. To understand why we need to make a quick detour in the problem of high dimensionality classification and discuss the \textit{curse of dimensionality}. Let us consider the unit cube \(\left[0,1\right]^{d}\) with \( d\) an integer and \( m\) points in it distributed randomly. How big should be the length \( l\) of the smallest hyper-cube to contain at least \( 1\) point? We can easily calculate it as
\begin{equation}
l^{d}\approx\frac{1}{m}\rightarrow l\approx\left(\frac{1}{m}\right)^{1/d}
\end{equation}
We can easily calculate this value of \( l\) for various values of \( d\). Let us suppose that we consider \( m = 1000\) and summarize the results in Table \ref{tab:res1}.

\begin{table}[hbt]
\begin{centering}
\begin{adjustbox}{max width=\textwidth}
\begin{tabular}{p{3.8cm}p{2.7cm}}
\multicolumn{1}{p{3.8cm}}{\centering
d} & 
\multicolumn{1}{p{2.7cm}}{\centering
l} \\ 
\hline
\multicolumn{1}{p{3.8cm}}{\centering
2} & 
\multicolumn{1}{p{2.7cm}}{\centering
0.003} \\ 
\multicolumn{1}{p{3.8cm}}{\centering
10} & 
\multicolumn{1}{p{2.7cm}}{\centering
0.50} \\ 
\multicolumn{1}{p{3.8cm}}{\centering
100} & 
\multicolumn{1}{p{2.7cm}}{\centering
0.93} \\ 
\multicolumn{1}{p{3.8cm}}{\centering
1000} & 
\multicolumn{1}{p{2.7cm}}{\centering
0.99} \\ 
\end{tabular}
\end{adjustbox}
\caption{Length \( l\) of the smallest hyper-cube to contain at least \( 1\) point from a population of randomly distributed \( m\) points.}\label{tab:res1}
\end{centering}
\end{table}

Furthermore, as you can see the data becomes so sparse in high dimensions that you need to consider the entire hyper cube to capture one single observation. When the data becomes so sparse the number of observations you will need to train an algorithm properly becomes much bigger than the size of existing datasets. 
We could look at this differently. Let us consider now a small hypercube of side \( l = 1/10\). How many observations we will find on average in this small portion of the hypercube? This is easy to calculate and is given by
\begin{equation}
\frac{m}{10^{d}}
\end{equation}
You can see that this number is very small for high values of \( d\). For example, if we consider \( d = 100\) is easy to see that we would need more observations than atoms in the universe\footnote{\url{https://www.universetoday.com/36302/atoms-in-the-universe/} } to find at least one observation in that small portion of the hypercube.
\begin{note}
\textbf{Note }Doing dimensionality reduction is a very viable method for reducing dramatically running time while incurring in an only small drop in accuracy. In high dimensionality datasets this becomes fundamental due to the curse of dimensionality.
\end{note}
\subsection{Anomaly Detection}

\begin{figure}[hbt]
\centering
\includegraphics[width=6.58cm,height=6.56cm]{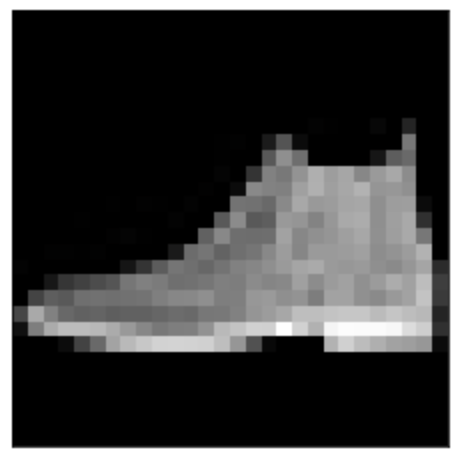}
\caption{one random image from the Zalando MNIST dataset.}\label{fig:anom1}
\end{figure}

Autoencoders are often used to perform anomaly detection on the most different datasets. The best way to understand how anomaly detection works with autoencoders is to look at it with a practical example. Let us consider an autoencoder with only three layers with 784 neurons in the first, 64 in the latent feature generation layer, and again 784 neurons in the output layers. We will train it with the MNIST dataset and in particular with the 60000 training portion of it as we have done in the previous sections of the article. Now let us consider the Fashion MNIST dataset. Let us choose an image of a shoe (see Figure \ref{fig:anom1}) from this dataset
and add it to the testing portion of the MNIST dataset. The original testing portion of MNIST has 10000 images. With the shoe we will have a 10001 images dataset. How can we use an autoencoder to find the shoe automatically in those 10001 images? Note that the shoe is an "outlier", an "anomaly" since it is an entirely different image class than hand-written digits. To do that we will take the autoencoder we trained with the 60000 MNIST images and with it we will calculate the reconstruction error for the 10001 test images.

\begin{figure}[hbt]
\centering
\includegraphics[width=12.61cm,height=7.02cm]{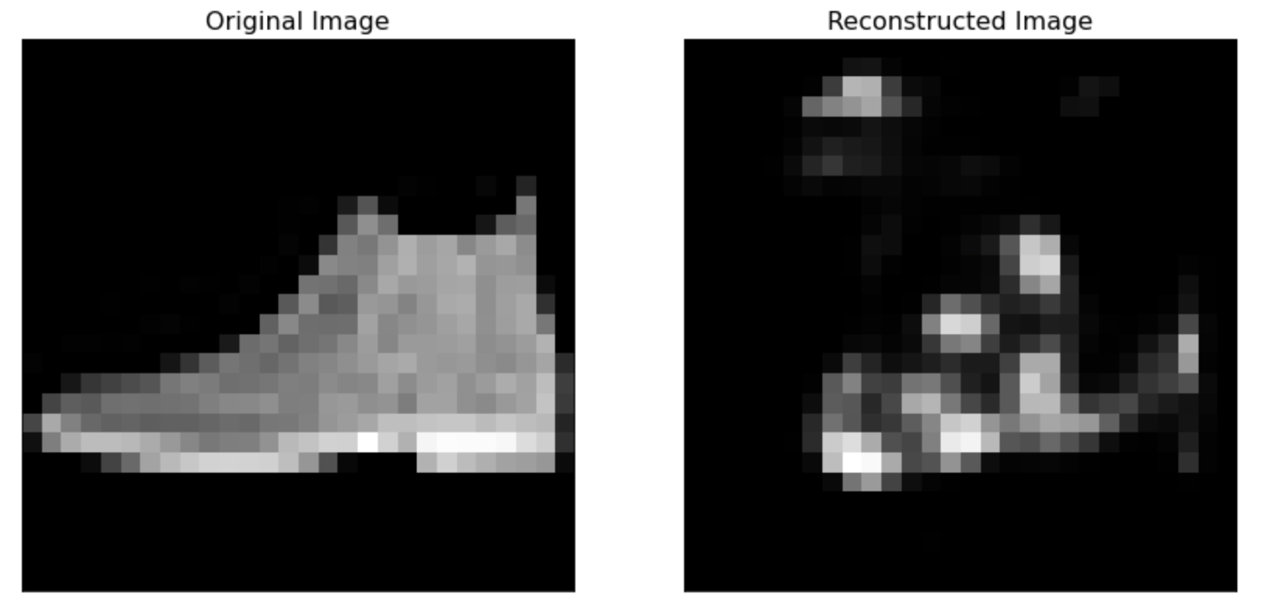}
\caption{The shoe and the autoencoder's reconstruction trained on the 60000 hand-written images of the MNIST dataset. This image has the biggest RE in the entire 10001 test dataset we built with a value of 0.062.}\label{fig:recon2}
\end{figure}

The main idea is that since the autoencoder has only seen hand-written digits images, it will not be able to reconstruct the shoe image. Therefore we expect this image to have the biggest reconstruction error. We can check if that is the case by taking the top 2 reconstruction errors. For this example, we have used the MSE for the reconstruction error. You can check the code of this example at \url{https://adl.toelt.ai}. The shoe has the highest reconstruction error: 0.062. The autoencoder is not able to reconstruct the image as it can be seen from Figure \ref{fig:recon2}.

\begin{figure}[hbt]
\centering
\includegraphics[width=12.61cm,height=7.09cm]{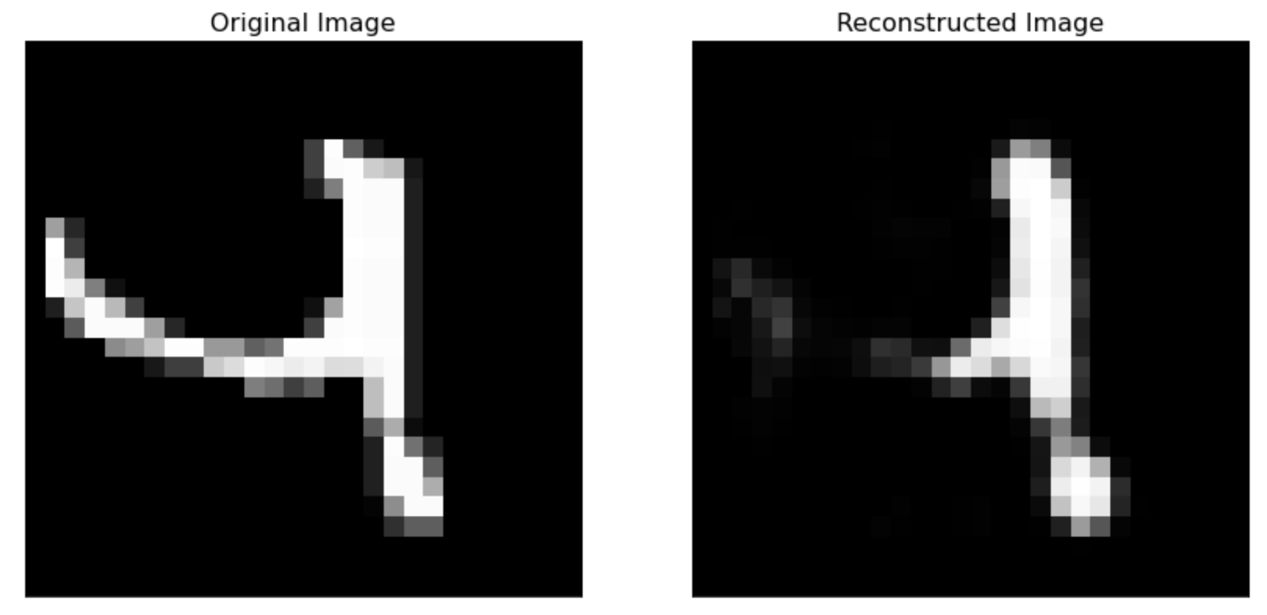}
\caption{the image with the second biggest RE in the 10001 test dataset: 0.022.}\label{fig:recon3}
\end{figure}

The second biggest RE is slightly less than one third of that of the shoe: 0.022, indicating that the autoencoder is doing quite a good job in understanding how to reconstruct hand-written digits. You can see the image with the second biggest RE in Figure \ref{fig:recon3}. This image could also be classified as an outlier, as is not completely clear if it is a 4 or an incomplete 9.
The readers with most experience may have noticed that we trained our autoencoders on a dataset without any outliers and applied it to a second dataset with outliers. This is not always possible as very often the outliers are not known and are lost in a big dataset. In general, one wants to find outliers in a single big dataset without any information on how many there are or how they look like. Generally speaking, anomaly detection can be done following the main steps below.
\begin{itemize}
\item One train an autoencoder on the entire dataset (or if possible, on a portion of the dataset known \textbf{not} to have any outlier).
\item
For each observation (or input) of the portion of the dataset known to have the wanted outliers one calculates the RE.
\item
One sorts the observations by the RE.
\item
One classifies the observations with the highest RE as outliers. Note that how many observations are outliers will depend on the problem at hand and require an analysis of the results and usually lot of knowledge of the data and the problem.
\end{itemize}
Note that if one train the autoencoder on the entire dataset at disposal, there is an essential assumption: the outliers are a negligible part of the dataset and their presence will not influence (or will influence in an insignificant way) how the autoencoder learns to reconstruct the observations. This is one of the reasons why regularization is so essential. If the autoencoders would learn the identity function, anomaly detection could not be done.
A classic example of anomaly detection is finding fraudulent credit card transactions (the outliers). This case usually presents ca. 0.1$\%$ fraudulent transactions and therefore this would be a case that would allow us to train the autoencoder on the entire dataset. Another is fault detection in industrial environment.
\begin{note}
\textbf{Note} If one train the autoencoder on the entire dataset at disposal, there is an essential assumption: the outliers are a negligible part of the dataset and their presence will not influence (or will influence in an insignificant way) how the autoencoder learns to reconstruct the observations.
\end{note}
\subsubsection{Model Stability – a short note}

Note that doing anomaly detection as described in the previous section seems easy, but those methods are prone to overfitting and give often inconsistent results. This means that training an autoencoder with a different architecture may well give different REs and therefore other outliers. There are several ways of solving this problem, but one of the simplest ways of dealing with instability of results is to train different models and then take the average of the REs. Another often used technique involves taking the maximum of the REs evaluated from several models. 
\begin{note}
\textbf{Note }Anomaly detection done with autoencoders is prone to problems as overfitting and unstable results. It is essential to be aware of these problems and check the results coming from different models to interpret the results correctly.
\end{note}
Note that this section serves to give you some pointers and is not meant to be an exhaustive overview on how to solve this problem.
Like autoencoders ensembles\footnote{ See for example \url{https://saketsathe.net/downloads/autoencode.pdf}.}, more advanced techniques are also used to deal with problems of instable results coming, for example, from small datasets.

\subsection{Denoising autoencoders}

\begin{figure}[hbt]
\centering
\includegraphics[width=12.16cm,height=11.92cm]{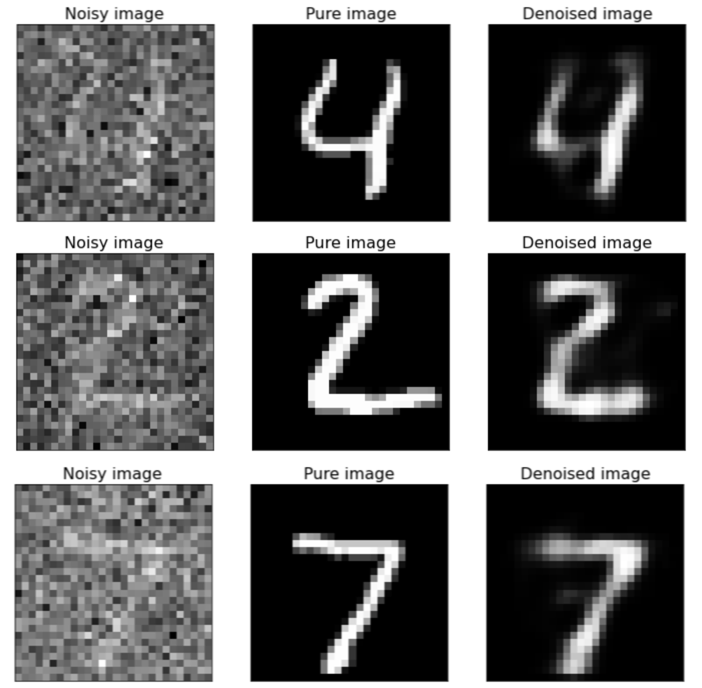}
\caption{Results of a denoising FFA autoencoder with 3 layers and 32 neurons in the middle layer. The noise has been generating adding a real number between 0 and 1 taken from a normal distribution. For details see the code at \url{https://adl.toelt.ai}.}\label{fig:recon5}
\end{figure}

Denoising autoencoders\footnote{ Vincent, P., Larochelle, H. Bengio, Y. Manzagol, P.A.: Extracting and composing robust features with denoising autoencoders. In: Proceedings of the 25th International Conference on Machine Learning, ICML ’08, pp. 1096-1103. ACM, New York, NY USA (2008)} are developed to auto-correct errors (noise) in the input observations. As an example, imagine the hand-written digits we considered before where we added some noise (for example Gaussian noise) in the form of changing randomly the gray values of the pixels. In this case the autoencoders should learn to reconstruct the image without the added noise. As a concrete example, consider the MNIST dataset. We can add to each pixel a random value generated by a normal distribution scaled by a factor (you can check the code at \url{https://adl.toelt.ai} in article 25). We can train an autoencoder using as input the noisy images, and as output the original images. The model should learn to remove the noise, since it is random in nature and have no relationship with the images. 
In Figure \ref{fig:recon5} you can see the results. In the left column you see the noisy images, in the middle the original ones and on the right the de-noised images. It is quite impressive how well it works. Figure \ref{fig:recon5} has been generated by training a FFA autoencoder with 3 layers and 32 neurons in the middle layer.

\section{Beyond FFA – autoencoders with convolutional layers}

\begin{figure}[hbt]
\centering
\includegraphics[width=12.61cm,height=7.3cm]{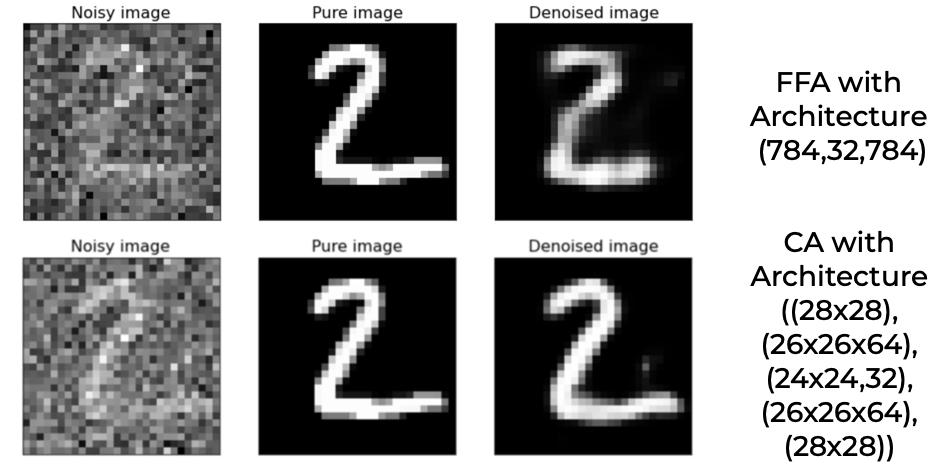}
\caption{Comparison of the results of a FAA (with architecture (784,32,784)) and of a Convolutional Autoencoder (CA) (with architecture ((28x28), (26x26x64), (24x24,32), (26x26x64), (28x28); keep in mind the layers are convolutions, so the first two numbers indicate the tensor dimensions and the third the number of kernels, with kernel size 3x3). The two autoencoders have been trained with the same parameters. You can check the code at \url{https://adl.toelt.ai}.}\label{fig:recon6}
\end{figure}

In this article we have described autoencoders with a feed-forward architecture. There is no rule and autoencoders with convolutional layers works as well, and often (especially when dealing with images) are much more efficient. For example, in Figure \ref{fig:recon6} you can see a comparison of the results of a FAA (with architecture (784,32,784)) and of a Convolutional Autoencoder (CA) (with architecture ((28x28), (26x26x64), (24x24,32), (26x26x64), (28x28); keep in mind the layers are convolutions, so the first two numbers indicate the tensor dimensions and the third the number of kernels, that in this example had a size of 3x3). The two autoencoders have been trained with the same parameters (epochs, mini-batch size, etc.). You can see how a CA gives better results than a FAA, since we are dealing with images. To be fair, note that the feature generating layer is only marginally smaller than the input layer in this example. The purpose of this example is only to show you how also convolutional autoencoders are a viable solution that works very well in many practical applications.

Another important aspect is that the feature generating layer could be a convolutional layer but could also be a dense one. There is not fix rule and testing is required to find the best architecture for your problem and how you want to model your latent features: as a tensor (multi-dimensional array) or as a one-dimensional array of real numbers.

\section{Code Examples}

On \url{https://adl.toelt.ai} you will find examples of autoencoders, anomaly detection with autoencoders and denoising with autoencoders as described in this article.

\section{Further Readings}

\begin{itemize}
\item Deep Learning Tutorial from Stanford University

\url{http://ufldl.stanford.edu/tutorial/unsupervised/Autoencoders/}

\item Building autoencoders in Keras

\url{https://blog.keras.io/building-autoencoders-in-keras.html}

\item Introduction to autoencoders in TensorFlow

\url{https://www.tensorflow.org/tutorials/generative/autoencoder}

\item Bank, D., Koenigstein, N., and Giryes, R., ``Autoencoders", arXiv e-prints, 2020,

\url{https://arxiv.org/abs/2003.05991}

\item R. Grosse, University of Toronto, Lecture on autoencoders

\url{http://www.cs.toronto.edu/~rgrosse/courses/csc321_2017/slides/lec20.pdf} 
\end{itemize}

\end{document}